\documentclass[conference]{IEEEtran}
\usepackage{cite}
\usepackage{amsmath,amssymb,amsfonts, amsthm}
\usepackage{graphicx}
\usepackage{textcomp}
\usepackage{xcolor}
\usepackage{amsthm}
\usepackage{algorithm}
\usepackage[noend]{algpseudocode}
\usepackage[colorlinks=true]{hyperref}
\usepackage{multirow}
\def\BibTeX{{\rm B\kern-.05em{\sc i\kern-.025em b}\kern-.08em
    T\kern-.1667em\lower.7ex\hbox{E}\kern-.125emX}}

\newtheorem{definition}{Definition}

\usepackage{mathtools}

\begin{document}

\title{Network Embedding: An Overview}

\author{\IEEEauthorblockN{Nino Arsov}
\IEEEauthorblockA{\textit{Faculty of Computer Science and Engineering} \\
\textit{Ss. Cyril and Methodius University}\\
Skopje, Macedonia \\
narsov@manu.edu.mk}
\and
\IEEEauthorblockN{Georgina Mirceva}
\IEEEauthorblockA{\textit{Faculty of Computer Science and Engineering} \\
\textit{Ss. Cyril and Methodius University}\\
Skopje, Macedonia \\
georgina.mirceva@finki.ukim.mk}
}

\maketitle

\begin{abstract}
Networks are one of the most powerful structures for modeling problems in the real world. Downstream machine learning tasks defined on networks have the potential to solve a variety of problems. With link prediction, for instance, one can predict whether two persons will become friends on a social network. Many machine learning algorithms, however, require that each input example is a real vector. Network embedding encompasses various methods for unsupervised, and sometimes supervised, learning of feature representations of nodes and links in a network. Typically, embedding methods are based on the assumption that the similarity between nodes in the network should be reflected in the learned feature representations. In this paper, we review significant contributions to network embedding in the last decade. In particular, we look at four methods: Spectral Clustering, \textsc{DeepWalk}, Large-scale Information Network Embedding (LINE), and node2vec. We describe each method and list its advantages and shortcomings. In addition, we give examples of real-world machine learning problems on networks in which the embedding is critical in order to maximize the predictive performance of the machine learning task. Finally, we take a look at research trends and state-of-the art methods in the research on network embedding.
\end{abstract}

\begin{IEEEkeywords}
networks, network embedding, unsupervised learning, latent feature representations
\end{IEEEkeywords}

\section{Introduction}
\label{sec:introduction}

Today, we live in a world that is connected more than ever before. But even before the rise of the internet and availability of vast amounts of data, the world was connected in many different ways, such as social and professional acquaintances and interactions. Today, however, centralized sources of data residing on digital storage systems reveal even more ways in which entities connect and interact with each other. Typical examples include social networks and scientific collaboration, among others. All these interactions can be easily represented with networks, known formally as graphs. The area dealing with this is called \textit{network science}. One of the best resources on network science is~\cite{barabasi2016network}. A graph is a discrete structure that can take many shapes and be of many different types. A graph consists of a set of nodes and links that connect pairs of nodes. Moreover, the links can be either directed or undirected and have weights that quantify the relationship between the pair of nodes they connect. It is a discrete structure studied in discrete mathematics. In general, many real-world scenarios can be modeled using networks.

In recent years, machine learning has taken network science to a different level where the focus is concentrated on prediction tasks. For instance, one may want to predict the existence of a link between a pair of nodes (also known as \textit{link prediction}). Another example is automatic detection of communities within networks (also known as \textit{community detection}). Furthermore, nodes and links in a network may have attributes that describe the entity represented with the particular node that a scientist would like to predict. All these are examples of \textit{downstream machine learning tasks}. Machine learning algorithms use real-valued input vectors and outputs to learn a latent function that maps each input vector into an output. In machine learning, inputs are either classified, i.e. assigned one or more labels from a finite set of labels, known as classes, or inputs are mapped to a real number that represents some quantity such as a product price or a custom measure. The former case is known as \textit{classification}, while the latter is referred to as a \textit{regression} task. In link prediction, for instance, two real-valued vectors that represent a pair of nodes are passed on to a machine learning algorithm to predict the existence of a link between them. This is a binary classification problem, in which the output label is either 0 or 1, which indicates the link's existence. Another task is node classification in which each node is assigned a class label, or even a real-number. The latter is known as node regression.

\textbf{Network embedding.} Network embedding refers to the approach of learning latent low-dimensional feature representations for the nodes or links in a network. The basic principle is to learn encodings for the nodes in the network such that the similarity in the embedding space reflects the similarity in the network.
The scope of node embedding is varying and applicable to all kinds of different graph types. The advantage of node embedding as a technique is that it does not require feature engineering by domain experts.

    

This paper is organized as follows: in Section~\ref{sec:preliminaries} we introduce various definitions and preliminaries to network embedding, in Section~\ref{sec:methods} we briefly review four network embedding methods, in Section~\ref{sec:case_studies} we describe three case studies of network embedding used in subsequent downstream tasks, borrowed from~\cite{grover2016node2vec}, in Section~\ref{sec:current_trends} we conclude the paper and talk about the latest state-of-the-art network embedding methods.

\section{Preliminaries}
\label{sec:preliminaries}
In this section, we formally give several definitions required to introduce node embedding techniques. First, we define graphs as a discrete structure. 

\begin{definition}{Graph.}\\
A graph $G(V, E)$ is a collection that amounts to a set of nodes $V=\{v_1,\ldots,v_n\}$, called \textit{nodes}, and a set of edges $E=\{e_{ij}\,|\,1\leq i,j\leq n\}\subseteq V\times V$, called \textit{links}. When $G$ is an undirected graph, if $e_{ij}\in E$, then $e_{ji}\in E$ and vice versa, and when $G$ is directed, $e_{ij}\in E$ does not necessarily imply that $e_{ji}\in E$.
\end{definition}

\begin{definition}{Weighted graph.}\\
A weighted graph $G(V,E,W)$ is a collection that amounts to a set of nodes $V=\{v_1,\ldots,v_n\}$, a set of links $E=\{e_{ij}\,|\,1\leq i,j\leq n\}\subseteq V\times V$, and a set of weights $W=\{w_{ij}\,|\, 1\leq i,j \leq n, w_{ij}\geq 0, w_{ij}\in \mathbb{R}_+\}.$ If $e_{ij}\notin E$, then $w_{ij}=0$, and otherwise $w_{ij}>0.$ 
\end{definition}

\noindent
The neighborhood of a node is generated using a search strategy that traverses the graph, such as Breadth-First Search (BFS), Depth-First Search (DFS), or a random walk. The neighborhoods of different nodes can have different sizes and they can overlap, i.e., $N(u)\cap N(v) \neq \varnothing.$

We now proceed with definitions related to node embedding. A node embedding is also known as a feature vector or a feature representation. The dimensionality of the embedding is given by $d\geq 1$ and is usually assumed to be known before the learning process takes place.

\begin{definition}{Node embedding.}\\
Let $d\geq 1$ be the dimensionality of the node embeddings. A node embedding function $f$ is a map $f:V \longrightarrow \mathbb{R}^d$ maps each node $v\in V$ to a real-valued feature vector in $\mathbb{R}^d.$ 
\end{definition}

\section{Selected Node Embedding Methods}
\label{sec:methods}

In this section we give a brief review of four prominent network embedding methods: \textit{Spectral Clustering}~\cite{tang2011leveraging}, \textit{\textsc{DeepWalk}}~\cite{perozzi2014deepwalk}, Large-scale Information Network Embedding (\textit{LINE})~\cite{tang2015line}, and node2vec~\cite{grover2016node2vec}. They all have the same goal of learning optimal node embeddings for large networks that can later be used in any downstream machine learning task.

\subsection{Spectral Clustering}
Spectral Clustering is a matrix factorization approach to network embedding based on the Laplacian matrix of a graph $G$. 
Spectral clustering was originally proposed to address the problem of partitioning a graph into disjoint sets. Here, the edges of a graph can have weights, denoting the similarity between nodes. Intuitively, we want to find a partition of the graph, so that the edges between groups have a small weight and the edges within a group have a large weight. This is closely related to the minimum-cut problem. For two disjoint node sets $B,C \subset V$, the cut between $B$ and $C$ is defined as

$$
Cut(B,C) = \sum_{v_i\in B, v_j \in C}A_{ij},
$$
where $A$ is the adjacency matrix of the graph. Any $d$-way partition  $(C_1, C_2, \ldots, C_d)$ should satisfy $\bigcup_{i=1}^d C_i=V,$ and $C_i \cap C_j=\varnothing$ for all $i\neq j$.

In~\cite{tang2011leveraging}, spectral clustering was chosen to extract node representations in social networks due to its effectiveness in various domains and the availability of a huge number of existing linear algebra packages to help solve the problem.

To find a good $k$-way partition, the problem can be formulated as $\min cut(C_1,C_2,\ldots, C_k)=\sum_{i=1}^k cut(C_i, V/C_i)$.

In practice, this formulation of the problem may return trivial partitions like a group consisting of only one node, separated from the rest of the network. There exist alternative objectives capable of finding a somehow ``balanced'' partitioning by additionally taking the group size into account~\cite{tang2011leveraging}. One commonly used objective is the normalized cut

$$
Ncut(C_1,\ldots, C_k)=\frac{1}{k}\sum_{i=1}^k\frac{cut(C_i,V/C_i)}{vol(C_i)},
$$
where $vol(C_i)=\sum_{v_j\in C_i}d_j$, and $d_j$ is the degree of node $v_j$.

Let 

\begin{equation}
H_{ij} = \begin{cases}
        1/\sqrt{vol(C_j)}\quad \text{if node } i \text{ belongs to community } C_j\\
        0\quad \text{otherwise}.
    \end{cases}
    \label{eq:asdasd}
\end{equation}
Then
$$
Ncut(C_1,C_2,\ldots, C_k)=\frac{1}{k}Tr(H^TLH),
$$
such that $L=D-A$ is the graph Laplacian. Considering that $H^TDH=I$, the Ncut problem can be rewritten as

\begin{flalign*}
& \min_{C_1,\ldots,C_k} Tr(H^TLH)\\
& \text{s.t. }  H^TDH = I\\
&H \text{ conforms to Equation (2)} 
\end{flalign*}

\noindent
If we define $S=D^{1/2}H$, the problem can be transformed to
\begin{flalign}
\min_{S} Tr(S^T\Tilde{L}S)\nonumber\\
\text{s.t. } S^TS=I,
\end{flalign}
where $\Tilde{L}=D^{-1/2}LD^{-1/2}=I-D^{-1/2}AD^{-1/2}$ is the normalized Laplacian~\cite{tang2011leveraging}.

The optimal solution of $S$ corresponds to the first $d$ eigenvectors of the normalized graph Laplacian $\Tilde{L}$ with the smallest eigenvalues. Typically in spectral clustering, a post-processing step like $k$-means clustering is applied to $S$ or $H$ to find a disjoint partition~\cite{tang2011leveraging}.

In summary, given a network $A$, spectral clustering is done by constructing a normalized Laplacian $\Tilde{L}$ and then computing the first (smallest) $d$ eigenvectors as the social dimensions, i.e., the feature representations of the nodes in the network. Finally, spectral clustering selects the $d$ smallest eigenvectors of the normalized Laplacian $\Tilde{L}$ as node embeddings~\cite{tang2011leveraging}.

\subsection{\textsc{DeepWalk}}
In \textsc{DeepWalk}, deep learning (unsupervised feature learning) was used for the first time to learn social representations of a graph's nodes by modeling a stream of short random walks. The algorithm learns latent feature representations that encode social relations in a continuous vector space with a relatively small number of dimensions. \textsc{DeepWalk} generalizes neural language models to process a special
language composed of a set of randomly-generated walks.
These neural language models have been used to capture the
semantic and syntactic structure of human language, and
even logical analogies~\cite{perozzi2014deepwalk}.

\textsc{DeepWalk} outperformed other latent representation methods for creating social dimensions, especially when the labelled nodes are scarce. The representations learned by \textsc{DeepWalk} make strong predictive performance with very simple linear models highly possible. In addition, the inferred representations are general and can be combined with any classification method~\cite{perozzi2014deepwalk}. \textsc{DeepWalk} is an online algorithm and is trivially parallelizable.

\textbf{Problem definition.} \textsc{DeepWalk} considers the problem of classifying members of a social network into one or more classes (categories). Let $G=(V,E)$ and let $G_L=(V, E, X, Y)$ be a partially labeled social network, with input features $X\in\mathbb{R}^{V\times S}$, where $S$ is the size of the feature space for each attribute vector, and $Y\in\mathbb{R}^{|V|\times |\mathcal{Y}|}$ given that $\mathcal{Y}$ is the set of possible labels.

The goal of \textsc{DeepWalk} is to learn $X_E \in \mathbb{R}^{|V|\times d}$, where $d$ is a small number of latent dimensions. These low-dimensional representations
are distributed; meaning each social phenomena is expressed
by a subset of the dimensions and each dimension contributes
to a subset of the social concepts expressed by the space~\cite{perozzi2014deepwalk}.

The method satisfies these requirements by learning representation for nodes from a stream of short random walks,
using optimization techniques originally designed for language modeling~\cite{perozzi2014deepwalk}.

\textbf{\textsc{DeepWalk}. }The algorithm consists of two main components; first a
random walk generator, and second, an update procedure.
The random walk generator takes a graph $G$ and samples
uniformly a random node $v_i$ as the root of the random
walk $\mathcal{W}_{v_i}$.
A walk samples uniformly from the neighbors
of the last node visited until the maximum length $(t)$ is
reached. While the length of the random walks in
the experiments is fixed, there is no restriction for the
random walks to be of the same length. These walks could
have restarts (i.e., a teleport probability of returning back
to their root), but the preliminary results have not shown any
advantage of using restarts. In practice, the implementation
specifies a number of random walks $\gamma$ of length $t$ to start at
each node~\cite{perozzi2014deepwalk}.

\textbf{SkipGram.} SkipGram is a language model that maximizes the co-occurrence probability among the words that appear within
a window, $w$, in a sentence. It approximates the conditional
probability using an independence assumption
as the following

$$
P(\{v_{i-w},\ldots, v_{i+w}\} \\ v_i|\Phi(v_i))= \prod_{j=i-w,\\j\neq i}^{i + w} P(v_j|\Phi(v_i)),
$$
where $\Phi(v_i)$ is a feature representation vector for node $v_i$. The purpose of the SkipGram model in \textsc{DeepWalk} is to capture the local structure around the node $v_i$, defined the surrounding $w$ nodes, i.e., neighbors.

\textbf{Hierarchical softmax.} Given that $u_k\in V$, calculating $P(u_k|\Phi(v_j))$ is not feasible. Computing the partition function to be used as a normalization factor is also expensive, and instead, \textsc{DeepWalk} resorts to the hierarchical softmax.  The nodes are assigned to the leaves of a binary tree, turning the prediction problem into maximizing the probability of a specific
path in the hierarchy. The process is shown in ~\cite[Fig. 3]{perozzi2014deepwalk}. This reduces the computational complexity of
calculating $P(u_k | \Phi(v_j))$ from 
$O(|V|)$ to $O(log|V|)$.

The training process can be sped up further by assigning shorter paths to the frequent nodes in the random walk. Huffman coding could help reduce the access time of frequent elements in the tree~\cite{perozzi2014deepwalk}.

Finally, the complete \textsc{DeepWalk} algorithm and SkipGram are given in~\cite{perozzi2014deepwalk}.



\subsection{LINE}
LINE (short for Large-Scale Information Network Embedding) is a representation learning algorithm that learns an embedding model for real world information networks.

In practice, information networks can be either directed (e.g., citation networks) or undirected (e.g., social network of users in Facebook). The weights of the edges can be either binary or take any real value. Embedding an information network into a low-dimensional space is useful in a variety of applications. To conduct the embedding, the network structures must be preserved. The first intuition is that the local network structure, i.e., the local pairwise proximity between the nodes, must be preserved~\cite{tang2015line}.

The first-order proximity between two nodes $u$ and $v$ is expressed by the edge weight $w_{uv}$. The second-order proximity is expressed by the similarity between $p_u$ and $p_v$, where $p_u=(w_{u,1},\ldots,w_{u,|V|})$.

The LINE embedding model preserves both first- and second-order proximities.

\textbf{LINE with First-order Proximity.} 
The first-order proximity refers to the local pairwise proximity between the nodes in the network. To model the
first-order proximity, for each undirected edge $(i, j)$, we define the joint probability between node $v_i$ and $v_j$ as follows:$p_1(v_i,v_j)=1/exp(-u_i^T \cdot u_j)$,
where $u_i \in \mathbb{R}^d$ is the low-dimensional vector representation
of node $v_i$. This defines a distribution $p(\cdot,\,\cdot)$ over the
space $V \times V$ , and its empirical probability can be defined
as $\hat{p}_1(i,j) = w_{ij}$, where 
$W = \sum_{(i,j)\in E}w_{ij}$. To preserve the $W(i,j)\in E$
first-order proximity, a straightforward way is to minimize the following objective function based on Kullback-Leibler (KL) divergence is used for $d(\cdot,\,\cdot)$:

\begin{equation}
O_1 = -\sum_{(i,j)\in E}w_{ij}\log p_1(v_i, v_j).
\label{eq:o1}
\end{equation}

\noindent
The first-order proximity is only applicable for undirected graphs.

\textbf{LINE with Second-order Proximity.} 

For each directed edge $(i, j)$, the probability of “context” $v_j$ generated by node $v_i$ is defined as:

\begin{equation}
    O_2 = \sum_{i\in V} \lambda_i d(\hat{p}_2(\cdot|v_i),p_2(\cdot|v_i))
    \label{eq:o2}
\end{equation}

\noindent
As the importance of the nodes in the network may be different, $\lambda_i$ is introduced in the objective function to represent the prestige of vertex $i$ in the network, which can be measured by the degree or estimated through algorithms such as
PageRank. The empirical distribution $\hat{p}_2(\cdot|v_i)$ is defined
as $\hat{p}_2(v_j|v_i) = \frac{w_{ij}}{d_i}$ , where $w_{ij}$ is the weight of the edge $(i,j)$ and $d_i$ is the out-degree of node $i$, i.e., $d_i=\sum_{k\in N(i)}w_{ik}$, $N(i)$ is the set of out-neighbors of $v_i$~\cite{tang2015line}. In the original paper, the authors set $\lambda_i=d_i$ and take advantage of KL-divergence as the distance function. Plugging KL-divergence into Equation~\eqref{eq:o2} and setting $\lambda_i=d_i$ yields $$O_2= -\sum_{(i,j)\in E} w_{ij}\log p_2(v_j|v_i).$$

Combined first-order and second-order proximity was left as future work~\cite{tang2015line}.

\textbf{Negative sampling (model optimization).} Optimizing objective $O_2$ is computationally expensive as it requires the summation over the entire set of nodes to calculate the conditional probability $p(\cdot|v_i)$. The authors address this issue by adopting the approach of negative sampling~\cite{mikolov2013distributed}, which samples multiple negative edges according to some noisy distributions for each edge $(i,j)$.
The final objective function is 

$$
O_2=\log \sigma(u'_j \cdot u_i) + \sum_{i=1}^K E_{v_n\sim P_n(v)}\left[\log \sigma(-u'_n \cdot u_i) \right],
$$
where $\sigma(x)=1/(1+exp(-x))$ is the sigmoid function and $K$ is the number of negative edges. Moreover, the first term models the observed edges while the second term models the negative edges drawn from the noise distribution~\cite{tang2015line}.

\subsection{node2vec}

node2vec is an algorithmic framework for learning continuous feature representations for nodes in networks. In node2vec, the goal is to learn a mapping of nodes to a low-dimensional space of features that maximizes the likelihood of preserving network neighborhoods of nodes. The authors have used flexible notions of a node's neighborhood and have designed a biased random walk procedure that sufficiently explores diverse neighborhoods. In its core, node2vec is a semi-supervised algorithm. The key characteristic of node2vec is its scalability as it scales to networks of millions of nodes~\cite{grover2016node2vec}.

\textbf{Problem definition.} Let $G=(V,E)$ be a given network. The node2vec framework is general and applies to any (un)directed, (un)weighted network~\cite{grover2016node2vec}. Let $f: V \rightarrow \mathbb{R}^d$ be the mapping function from nodes to feature representations to be learned for future downstream tasks. Equivalently, $f$ is a matrix of size $|V|\times d$ parameters. For every source node $u\in V$, the \textit{network neighborhood}  of node $u$ generated through a neighborhood sampling strategy $S$ is defined $N_S(u)\subset V$. The authors then extend the SkipGram architecture to networks.

\noindent
In general, node2vec seeks to optimize an objective function that maximizes the log-probability of observing a network neighborhood $N_S(u)$ for a node $u$, conditioned on its feature representation, given by $f$:

\begin{equation}
\max_{f} \sum_{u\in V} \log P(N_S(u)|f(u)).
\label{eq:node2vec_o1}
\end{equation}

Since the problem given in the form above is intractable, the authors make two critical assumptions to make it tractable: conditional independence (the likelihood of observing a neighborhood node is independent of the likelihood of observing any other neighborhood node) and symmetry in the feature space (a source node $u$ and any neighborhood node $n_i\in N_S(u)$ have a symmetric effect over each other).
The likelihood is modeled using the softmax: 
    
    $
    P(n_i|f(u))=exp(f(n_i)\cdot f(u))/\sum_{v\in V}exp(f(v) \cdot f(u).
    $

The objective in Equation~\eqref{eq:node2vec_o1} simplifies to 

\begin{equation}
    \max_f \sum_{u \in V} \left[ -\log Z_u + \sum_{n_i \in N_S(u)} f(n_i)\cdot f(u)\right],
    \label{eq:node2vec_o2}
\end{equation}
where $Z_u=\sum_{v\in V}exp(f(u)\cdot f(v))$ is the partition function for node $u$. Unfortunately, computing $Z_u$ is computationally expensive for large networks, and therefore, following the LINE principles, $Z_u$ is approximated with negative sampling~\cite{grover2016node2vec}. The authors use stochastic gradient descent (SGD) to optimize Equation~\eqref{eq:node2vec_o2} and attain scalability for large networks.

\textbf{Neighborhood generating strategies.} The neighborhoods generated within node2vec are not restricted to just the immediate neighbors: the authors propose a randomized sampling strategy based on random walks to generate a diverse neighborhood~\cite{grover2016node2vec}. The neighborhood size is constrained to $k$. BFS and DFS are two extreme sampling strategies: the first samples the immediate neighbors of a node $u$, while the latter samples nodes sequentially at increasing distances from the source node $u$. Nodes sampled with these two strategies are capable of conforming to the homophily hypothesis~\cite{fortunato2010community,yang2014overlapping} and structural equivalence~\cite{hoff2002latent}. More information can be found in the original paper~\cite{grover2016node2vec}. Two parameters, $p$ and $q$, are used to control the random walks. First, $p$ controls the likelihood of immediately revisiting a node in the walk. Setting it to a high value makes the walk less incline to sample an already-visited node in the previous two steps. The parameter $q$ differentiates between ``inward'' and ``outward'' nodes. If $q<1$, the walk is more inclined to visit nodes further from $t$. This behavior is reflective of DFS. If $q>1$, the random walk is biased towards nodes close to node $t$, which is reflective of BFS.

Finally, node2vec initializes $r$ random walks per node to generate diverse sets of node neighborhoods. More details on the search strategy can be found in~\cite{grover2016node2vec}.
The node2vec algorithm is given in Algorithm~\ref{alg:node2vec}, and the biased random walk procedure is given in~\cite{grover2016node2vec}.

\begin{algorithm}
\caption{\textsc{node2vec}}
\label{alg:node2vec}
\begin{algorithmic}[1]
\Statex \textbf{LearnFeatures} (Graph $G=(V,E,W)$, Dimensions $d$, Walks per node $r$, Walk length $l$, Context size $k$, Return $p$, In-out $q$)
\State $\pi=\text{PreprocessModifiedWeights}(G, p, q)$
\State $G'=(V, E, \pi)$
\State Initialize $walks$ to Empty
\For{$iter=1,\ldots, r$}
\For{\textbf{all} nodes $u\in V$}
\State $walk=\text{node2vecWalk}(G',u,l)$
\State Append $walk$ to $walks$
\EndFor
\EndFor
\State $f=\text{StochasticGradientDescent}(k, d, walks)$
\State \textbf{return} $f$
\end{algorithmic}
\end{algorithm}
 The three phases of node2vec, i.e., preprocessing to compute transition probabilities, random walk simulations and optimization using SGD, are executed sequentially. Each phase is parallelizable and executed asynchronously, contributing to the overall scalability of node2vec.

\section{Case Studies}
\label{sec:case_studies}

\subsection{Case Study 1: Les Mis\`{e}rables Network}
\label{sec:usecases_1}

This section was borrowed from~\cite{grover2016node2vec}. It studies the Les Mis\`{e}rables network and four node embedding algorithms: Spectral Clustering, \textsc{DeepWalk}, LINE, and node2vec.

In this network, nodes correspond to characters in the novel Les Mis\`{e}rables~\cite{knuth1994stanford} and edges connect coappearing characters. The network has 77 nodes and 254 edges. The embedding dimension was set to $d = 16$ and node2vec was run to learn a feature representation for every node in the network. The feature representations are clustered using $k$-means and the nodes were colored according to the cluster assignments~\cite{grover2016node2vec}.
Fig.~\ref{fig:les_miserables} (top) shows the example when $p = 1, q = 0.5$. Regions of the network are colored using the same color. In this setting node2vec discovers clusters/communities of characters that frequently interact with each other in the major sub-plots of the novel. This characterization closely relates with homophily~\cite{grover2016node2vec}.
In order to discover which nodes have the same structural roles the authors use the same network but set $p = 1, q = 2$. Here, node2vec obtains a complementary assignment of node to clusters such that the colors correspond to structural equivalence as illustrated in Fig.~\ref{fig:les_miserables}(bottom). For instance, node2vec embeds blue-colored nodes close together. These nodes represent characters that act as bridges between different sub-plots of the novel. Similarly, the yellow nodes mostly represent characters that are at the periphery and have limited interactions. In~\cite{Figure 1}{grover2016node2vec}, the the top plot reflects homophily and the bottom plot represents structural equivalence.

\subsection{Case Study 2: Multilabel Classification}
\label{sec:usecases_2}

This section was also borrowed from~\cite{grover2016node2vec}. The authors compare the Macro-$F_1$ scores for multi-label classification on three datasets: BlogCatalog, Protein-Protein Interactions (PPI), and Wikipedia (see~\cite{grover2016node2vec}).

All these networks exhibit a fair mix of homophilic and structural equivalences~\cite{grover2016node2vec}.

The node feature representations were input to a one-vs-rest logistic regression classifier with $L_2$ regularization. The train and test data was split equally over 10 random instances. The authors used the Macro-$F_1$ scores for comparing performance, shown in Table~\ref{tbl:results}.

\begin{table}[]
\label{tbl:results}
\caption{Macro-$F_1$ Scores of different network embedding algorithms. This table is borrowed from the original node2vec paper~\cite{grover2016node2vec}.}
\begin{tabular}{r|ccc}
\multicolumn{1}{c|}{\multirow{2}{*}{\textbf{Algorithm}}} & \multicolumn{3}{c}{\textbf{Dataset}}                                                          \\
\multicolumn{1}{c|}{}                                    & \multicolumn{1}{c|}{BlogCatalog}     & \multicolumn{1}{c|}{PPI}             & Wikipedia       \\ \hline
Spectral Clustering                                                 & \multicolumn{1}{c|}{0.0405}          & \multicolumn{1}{c|}{0.0681}          & 0.0395          \\
\hline
\textsc{DeepWalk}                                                 & \multicolumn{1}{c|}{0.2110}          & \multicolumn{1}{c|}{0.1768}          & 0.1274          \\ \hline
LINE                                                     & \multicolumn{1}{c|}{0.0784}          & \multicolumn{1}{c|}{0.1447}          & 0.1164          \\ \hline
node2vec                                        & \multicolumn{1}{c|}{\textbf{0.2581}} & \multicolumn{1}{c|}{\textbf{0.1791}} & \textbf{0.1552} \\ 
node2vec settings ($p,q$)                                & \multicolumn{1}{c|}{(0.25, 0.25)}      & \multicolumn{1}{c|}{(4,1)}             & (4, 0.5)    \vspace{2mm}     
\end{tabular}
\end{table}



\subsection{Case Study 3: Link Prediction}
\label{sec:usecases_3}

In link prediction, the task is to predict the existence of links between pairs of nodes given a network with a certain fraction of edges removed.
Interestingly, the authors of node2vec were the first to use the learned feature representations for link prediction. In this task, the challenge is to combine the features for each pair of nodes; usually, a pair-wise operator between $f(u)$ and $f(v)$, such as the average ($\frac{f(u)+f(v)}{2}$), the Hadamard, i.e., pair-wise product ($(f(u)\circ f(v))_i=(f(u))_i(f(v))_i$), the Weighted-L1, and Weighted-L2 norm is used to combine the features.

The link prediction performance of Spectral Clustering, \textsc{DeepWalk}, LINE, and node2vec was tested on two additional datasets along PPI: the Facebook dataset~\cite{snap_datasets}, in which nodes represent users and edges represent friendship relation between any two users, and the arXiv ASTRO-PH dataset~\cite{snap_datasets}, a collaboration network generated from papers submitted to arXiv where nodes represent scientists and edges represent collaboration, that is, an edge is present between two scientists if they have collaborated in a paper. With respect to Area Under Curve (AUC) scores on link prediction, node2vec again outperforms both \textsc{DeepWalk} and LINE with gain up to 3.8\% and 6.5\%, respectively. The scores are summarized in Table~\ref{tbl:results_2} (borrowed from~\cite{grover2016node2vec}). The authors reported that the Hadamard product is the most stable and gives the best performance across all networks on average~\cite{grover2016node2vec}.

\begin{table}
\label{tbl:results_2}
\caption{AUC Scores for Link Prediction of the Four Algorithms Using Different Operators to Embed Links: (a) Average, (b) Hadamard, (c) Weighted-L1 Norm, and (d) Weighted-L2 Norm, borrowed from~\cite{grover2016node2vec}}
\begin{tabular}{c|rccc}
\multicolumn{1}{c|}{\multirow{2}{*}{\textbf{Operator}}} & \multicolumn{1}{c|}{\multirow{2}{*}{\textbf{Algorithm}}} & \multicolumn{3}{c}{\textbf{Dataset}}                \\
\multicolumn{1}{c|}{}                                   & \multicolumn{1}{c|}{}                                    & Facebook        & PPI             & arXiv           \\ \hline
\multirow{4}{*}{(a)}                                    & \multicolumn{1}{r|}{Spectral Clustering}                & 0.5960          & 0.6588          & 0.5812          \\
                                                        & \multicolumn{1}{r|}{DeepWalk}                           & 0.7238          & 0.6923          & 0.7066          \\
                                                        & \multicolumn{1}{r|}{LINE}                               & 0.7029          & 0.6330          & 0.6516          \\
                                                        & \multicolumn{1}{r|}{node2vec}                  & 0.7266          & 0.7543          & 0.7221          \\ \hline
\multirow{4}{*}{(b)}                                    & \multicolumn{1}{r|}{Spectral Clustering}                & 0.6192          & 0.4920          & 0.5740          \\
                                                        & \multicolumn{1}{r|}{DeepWalk}                           & \textbf{0.9680} & 0.7441          & 0.9340          \\
                                                        & \multicolumn{1}{r|}{LINE}                               & 0.9490          & 0.7249          & 0.8902          \\
                                                        & \multicolumn{1}{r|}{node2vec}                  & \textbf{0.9680} & \textbf{0.7719} & \textbf{0.9366} \\ \hline
\multirow{4}{*}{(c)}                                    & \multicolumn{1}{r|}{Spectral Clustering}                & 0.7200          & 0.6356          & 0.7099          \\
                                                        & \multicolumn{1}{r|}{DeepWalk}                           & 0.9574          & 0.6026          & 0.8282          \\
                                                        & \multicolumn{1}{r|}{LINE}                               & 0.9483          & 0.7024          & 0.8809          \\
                                                        & \multicolumn{1}{r|}{node2vec}                  & 0.9602          & 0.6292          & 0.8468          \\ \hline
\multirow{4}{*}{(d)}                                    & \multicolumn{1}{r|}{Spectral Clustering}                & 0.7107          & 0.6026          & 0.6765          \\
                                                        & \multicolumn{1}{r|}{DeepWalk}                           & 0.9584          & 0.6118          & 0.8305          \\
                                                        & \multicolumn{1}{r|}{LINE}                               & 0.9460          & 0.7106          & 0.8862          \\
                                                        & \multicolumn{1}{r|}{node2vec}                  & 0.9606          & 0.6236          & 0.8477          \vspace{2mm}
\end{tabular}
\end{table}

\section{Current Trends in Node Embedding Research}
\label{sec:current_trends}

The four approaches that we took a look at in this paper form the base of network embedding. The most efficient method is node2vec~\cite{grover2016node2vec} that easily outperforms the remaining three. During the last two years, however, there have been significant advances in developing novel embedding approaches applicable to various types of networks. The methods surveyed here are applicable to homogeneous networks, i.e., networks in which all nodes represent instances of the same entity. Network embedding in heterogeneous networks is more challenging and one of the methods that does this is metapath2vec~\cite{dong2017metapath2vec}. Next, Modulized Non-Negative Matrix Factorization (N-NMF) learns representations that preserve the communities within the network~\cite{wang2017community}. For networks in which the nodes have multiple attributes (also known as attributed networks), one can use label-informed attributed network embedding~\cite{huang2017label}. A framework called struc2vec~\cite{ribeiro2017struc2vec} learns embeddings that preserve the structural identity, which is a concept of symmetry in which network nodes are identified according 
to the network structure and their relationship to other nodes in a network. Nodes residing in different parts of a graph can have similar structural roles within their local 
network topology. This kind of embeddings can be learned via diffusion wavelets~\cite{donnat2018learning}. Node embedding can be extended to links that represent relationships in social networks, for instance---this is known as relationship embedding~\cite{lai2019transconv}. The following papers provide extensive surveys on network embedding:~\cite{goyal2018graph}, ~\cite{cui2018survey}.

\section{Conclusion}
\label{sec:conclusion}

Network embedding is critical for applying machine learning approaches that are becoming ubiquitous in network science. In this paper, we reviewed four important network embedding techniques: spectral clustering, \textsc{DeepWalk}, LINE, and node2vec. The representations learned by node2vec manifest the best performance in downstream tasks. Learning disentangled representations is a popular research trend that tries to bring network embedding to a new level in which the black-box model is replaced by methods that learn representations in which each original feature is represented by one or more dimensions in the learned embeddings. Network embedding methods have contributed to a large extent in applying machine learning in network science.

\bibliographystyle{IEEEtran}
\bibliography{IEEEabrv,refs.bib}

\end{document}